\documentclass{article} 
\usepackage{iclr2026_conference}
\usepackage{times}

\usepackage{amsmath,amsfonts,bm}

\def\eqref#1{equation~\ref{#1}}

\def\1{\bm{1}}

\DeclareMathAlphabet{\mathsfit}{\encodingdefault}{\sfdefault}{m}{sl}
\SetMathAlphabet{\mathsfit}{bold}{\encodingdefault}{\sfdefault}{bx}{n}

\usepackage{hyperref}[url]
\usepackage{url}
\usepackage{newfloat}
\usepackage{listings}
\usepackage{helvet}
\usepackage{courier}
\usepackage{graphicx}
\usepackage{caption}
\usepackage{natbib}
\usepackage{algorithm}
\usepackage{algorithmic}
\usepackage{booktabs}
\usepackage{multirow} 
\usepackage{array}
\usepackage{amssymb}
\usepackage{pifont}
\usepackage{wrapfig}

\title{UrbanFeel: A Comprehensive Benchmark for Temporal and Perceptual Understanding of City Scenes through Human Perspective}

\newcounter{example} 
\author{
  Jun He\textsuperscript{\rm 1},
  Yi Lin\textsuperscript{\rm 1},
  Zilong Huang\textsuperscript{\rm 1},
  Jiacong Yin\textsuperscript{\rm 1},
  Junyan Ye\textsuperscript{\rm 1}, 
  Yuchuan Zhou\textsuperscript{\rm 1}, \\
  \textbf{
  Weijia Li\textsuperscript{\rm 1}\thanks{ Corresponding author(s). E-mail(s): \texttt{\{liweij29, zhangx795\}@mail.sysu.edu.cn}},
  Xiang Zhang\textsuperscript{\rm 1}\footnotemark[\value{footnote}]
  } \\
  ~\textsuperscript{\rm 1}Sun Yat-sen University
}

\iclrfinalcopy

\begin{document}

\maketitle

\begin{abstract}
Urban development impacts over half of the global population, making human-centered understanding of its structural and perceptual changes essential for sustainable development. While Multimodal Large Language Models (MLLMs) have shown remarkable capabilities across various domains, existing benchmarks that explore their performance in urban environments remain limited, lacking systematic exploration of temporal evolution and subjective perception of urban environment that aligns with human perception. To address these limitations, we propose UrbanFeel, a comprehensive benchmark designed to evaluate the performance of MLLMs in urban development understanding and subjective environmental perception. UrbanFeel comprises 14.3K carefully constructed visual questions spanning three cognitively progressive dimensions: Static Scene Perception, Temporal Change Understanding, and Subjective Environmental Perception. We collect multi-temporal single-view and panoramic street-view images from 11 representative cities worldwide, and generate high-quality question-answer pairs through a hybrid pipeline of spatial clustering, rule-based generation, model-assisted prompting, and manual annotation. Through extensive evaluation of 20 state-of-the-art MLLMs, we observe that Gemini-2.5 Pro achieves the best overall performance, with its accuracy approaching human expert levels and narrowing the average gap to just 1.5\%. Most models perform well on tasks grounded in scene understanding. In particular, some models even surpass human annotators in pixel-level change detection. However, performance drops notably in tasks requiring temporal reasoning over urban development. Additionally, in the subjective perception dimension, several models reach human-level or even higher consistency in evaluating dimension such as beautiful and safety. Our results suggest that MLLMs are demonstrating rudimentary emotion understanding capabilities. Our UrbanFeel benchmark will be made publicly available.
\end{abstract}

\section{Introduction}

With over half of the global population now living in urban areas~\citep{worldbank_urban_2024}, understanding the dynamics of urban development has become increasingly critical for designing sustainable governance strategies, guiding urban policy, and promoting human-centric smart cities~\citep{yuan2024fusu, van2021multi,zhang2024predicting,li2023omnicity}. Compared to satellite imagery, which provides macro-scale, top-down observations, street-view imagery offers fine-grained, street-level perspectives that are more aligned with human visual perception~\citep{biljecki2021street, naik2017computer, wang2025exploring}. This unique characteristic enables it to capture subtle environmental changes within cities, making it a valuable data source for analyzing intra-urban transformation.

Recent research has explored the use of deep learning models in conjunction with street-view imagery to assess urban development stages~\citep{zhang2018measuring, alpherts2025emplace, ye2024cross}, visual quality~\citep{ito2024understanding, benidir2025change,ye2024leveraging}, and perceived livability~\citep{dubey2016placepulse, yang2024virl,li2025fine, ye2024sg}. However, these approaches face challenges in terms of generalization across modalities and cities. More importantly, they struggle to effectively quantify and interpret human subjective perception—an essential component of real-world urban understanding.

The advent of Large Language Models (LLMs) and Multimodal LLMs (MLLMs) has introduced new possibilities for tackling these limitations~\citep{Zhang2024UrbanSP,Xuan2025DynamicVLBM, ye2025satellite}. By leveraging massive amounts of multimodal pretraining data, MLLMs exhibit strong capabilities in spatial reasoning, visual-linguistic alignment, and commonsense inference. Initial attempts have applied these models to urban imagery tasks, such as vehicle trajectory prediction~\citep{liu2025citylens, lai2025ustbench} or scene understanding~\citep{yan2024urbanclip, feng2025urbanllava, feng2025citygpt, ye2024where}, and several early benchmarks have emerged to evaluate their performance on objective tasks such as image geolocalization~\citep{zhou2025urbench} and infrastructure inference~\citep{feng2025citybench}.

 \begin{figure}[t!]
     \centering
     \includegraphics[width=0.82\linewidth]{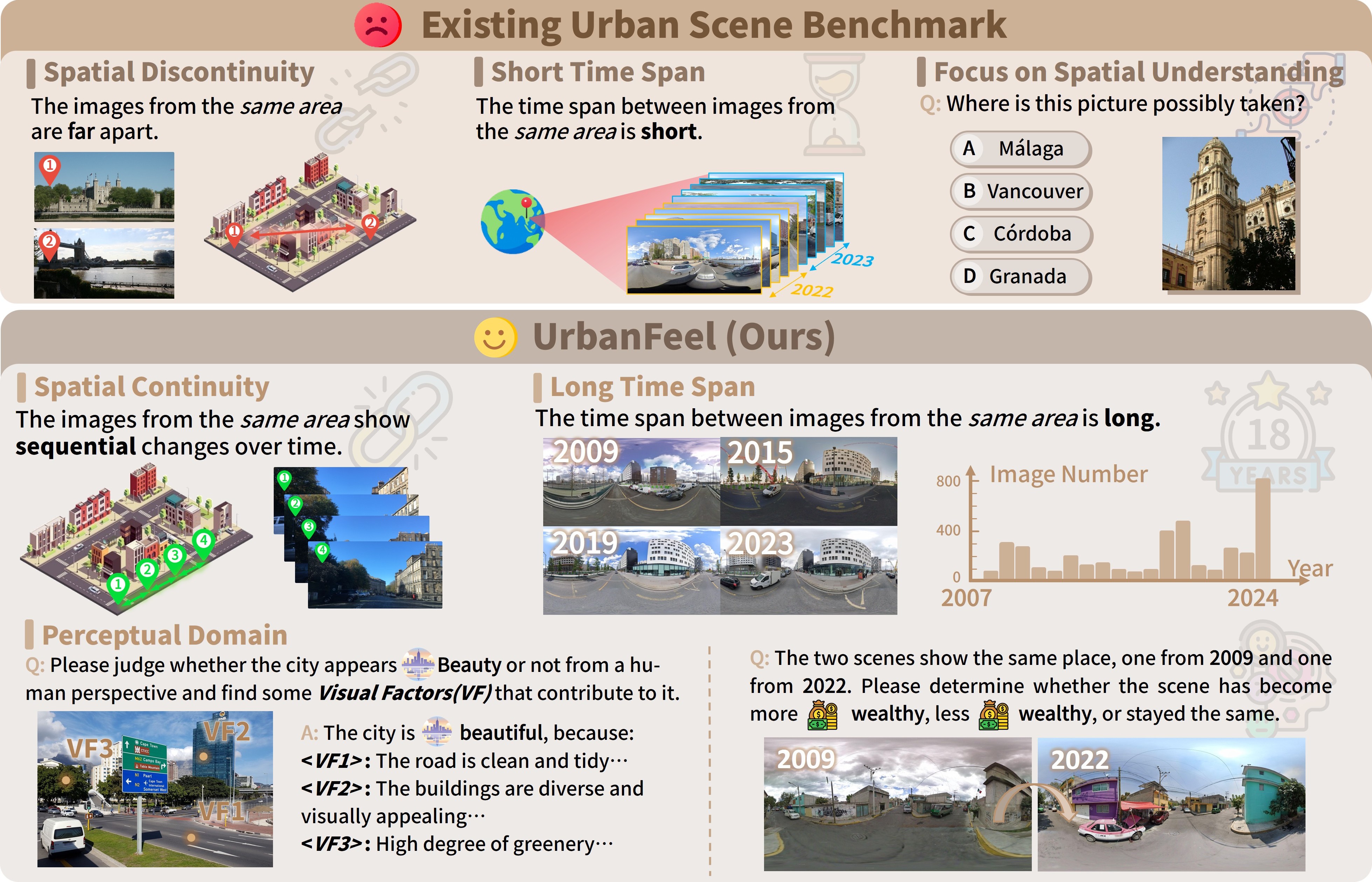}
     \vspace{-2mm}
     \caption{\textcolor{black}{Comparison with existing urban scene benchmarks. UrbanFeel introduces three key innovations: (1) spatially continuous street-view data that includes both single-view and panoramic imagery, (2) long-term temporal coverage spanning over 15 years, and (3) a novel evaluation dimension focused on subjective human perception (e.g., safety, beauty), enabling human-centered assessment beyond conventional spatial understanding.}}

     \label{fig:comparison}
     \vspace{-6mm}
 \end{figure}

\textcolor{black}{Prior work has largely been confined to static snapshots, focusing on objective recognition tasks such as autonomous driving or urban planning, while overlooking the historical dynamics of cities and thus failing to capture trajectories of development, renewal, and transformation. At the same time, physical changes in the built environment—such as renovation or decay—often reshape human perceptions of safety, beauty, and liveliness. However, existing benchmarks rarely examine how these perceptual shifts are linked to temporal urban evolution, leaving a critical gap in understanding the interaction between physical change and human experience.}

To bridge these gaps, we present \textbf{UrbanFeel}, a novel human-centric benchmark for evaluating MLLMs in the context of urban change perception. UrbanFeel defines 11 tasks across three dimensions—\textit{static scene perception}, \textit{temporal change understanding}, and \textit{subjective environmental perception}—to assess models' capabilities in recognition, reasoning, 0and alignment with human perception. Our benchmark emphasizes multi-view integration, temporal-spatial consistency, and perceptual alignment, aiming to push the boundaries of MLLMs toward more human-aligned urban understanding.

Our main contributions are summarized as follows:

\begin{itemize}
    \item We introduce \textbf{UrbanFeel}, a multi-perspective, multi-dimensional benchmark designed to evaluate MLLMs' performance on tasks related to urban development and human perception. UrbanFeel carefully designs 11 subtasks, focusing on evaluating the model's perception and understanding capabilities in three dimensions: Static Scene Perception, Temporal Change Understanding, and Subjective Environmental Perception.

    \item We design a scalable and interpretable task-querying framework, incorporating a diverse range of evaluation formats including binary classification, multiple-choice, sorting, and open-ended reasoning. To enhance explainability, we introduced manual annotation based on local visual evidence into the benchmark management process.

    \item We conduct a comprehensive evaluation of 20 state-of-the-art MLLMs on UrbanFeel, quantifying model differences across task categories and revealing that current models still fall short of human-level performance in spatial reasoning and subjective perception within urban change scenarios.

\end{itemize}

\section{Related Work}

\subsection{Urban Change and Perceptual Assessment}

With the accelerating pace of global urbanization, cities have undergone profound spatial and environmental transformations, prompting growing research interest in urban evolution~\citep{pandey2015urbanization,hatab2019urban,follmann2021city}. In recent years, street-view imagery has emerged as a valuable data source for urban change detection due to its close alignment with human perspectives~\citep{biljecki2021street, ito2024understanding}. For instance, Street2Vec embeds street-view images from different time points into a latent space to measure changes in urban physical structures~\citep{stalder2024street2vec}, while CityPulse~\citep{huang2024citypulse} constructs a long-term image sequence with semantic labels to detect binary changes in the built environment. Meanwhile, increasing attention has been paid to the subjective perception of urban environments~\citep{zhang2018measuring, fan2023urban}. The Place Pulse dataset~\citep{dubey2016placepulse} introduced perception dimensions such as beauty, wealth, and safety, and ChangeScore~\citep{naik2017computer} aligned semantic differences with perceptual shifts. However, existing studies often struggle with generalization across different cities and modalities, limiting their applicability in diverse urban environments. Moreover, few methods attempt to quantify or explain subjective human perception in a consistent and scalable way.

\subsection{Multimodel Large Language Models}

In recent years, Multimodal Large Language Models (MLLMs) such as Qwen~\citep{bai2023qwen}, GPT-4o~\citep{openai_gpt4o_2024}, and Gemini-2.5-pro~\citep{comanici2025gemini25} have achieved remarkable progress in image generation~\citep{anonymous2025icg}, visual reasoning~\citep{ZhangPWLSCMY25}, and cross-modal alignment~\citep{Wu2024SemanticAF,ye2024loki}. Leveraging large-scale pretraining and instruction tuning, these models show strong generalization in open-domain visual understanding~\citep{li2024llava, wu2024deepseek}, though their performance on domain-specific applications remains limited. In urban contexts, recent studies explored MLLMs’ zero-shot spatial reasoning. UrbanCLIP~\citep{yan2024urbanclip} aligns imagery with textual semantics via contrastive learning, while UrbanLLaVA~\citep{feng2025urbanllava} integrates street-view, structured data, and trajectories, achieving strong generalization on UBench. Despite these advances, a systematic framework for evaluating MLLMs on subjective urban perception or urban change assessment remains lacking. Prior efforts, such as~\citep{zhang2025urbansafety}, focus on isolated dimensions like safety, without addressing temporal coherence or perceptual consistency. Overall, existing work emphasizes static reasoning or functional classification, overlooking human-centric perceptual responses and their evolution over time.

\subsection{MLLM Benchmarks in Urban Scene}

With Multimodal Large Language Models (MLLMs) advance in image understanding~\citep{Ma2024GromaLV} and cross-modal reasoning~\citep{HuangCXL24a}, benchmark datasets have evolved accordingly. Early benchmarks centered on basic Visual Question Answering (VQA), but such tasks no longer capture the full potential of modern MLLMs. To address this, several expert-level benchmarks have been introduced for domain-specific tasks with greater semantic and spatial complexity, especially in urban contexts. For example, V-IRL~\citep{yang2024virl} focuses on street-view navigation and recognition; CityBench~\citep{feng2025citybench} targets urban identity and navigation, though with limited task diversity. UrBench~\citep{zhou2025urbench} incorporates multi-view imagery from street and remote sensing sources for spatial reasoning. CityLens~\citep{liu2025citylens} evaluates urban function modeling using socio-economic indicators, and USTBench~\citep{lai2025ustbench} assesses spatial planning via traffic and road network data. Despite these advances in modeling objective urban scenarios, most existing benchmarks are limited to static snapshots in time. They lack a comprehensive evaluation of models’ ability to capture the spatiotemporal evolution of urban environments, particularly how physical transformations affect human subjective perception responses.

\section{UrbanFeel}
\label{sec:urbanfeel}

\subsection{Overview}

\begin{figure}[t!]
    \centering
    \includegraphics[width=\linewidth]{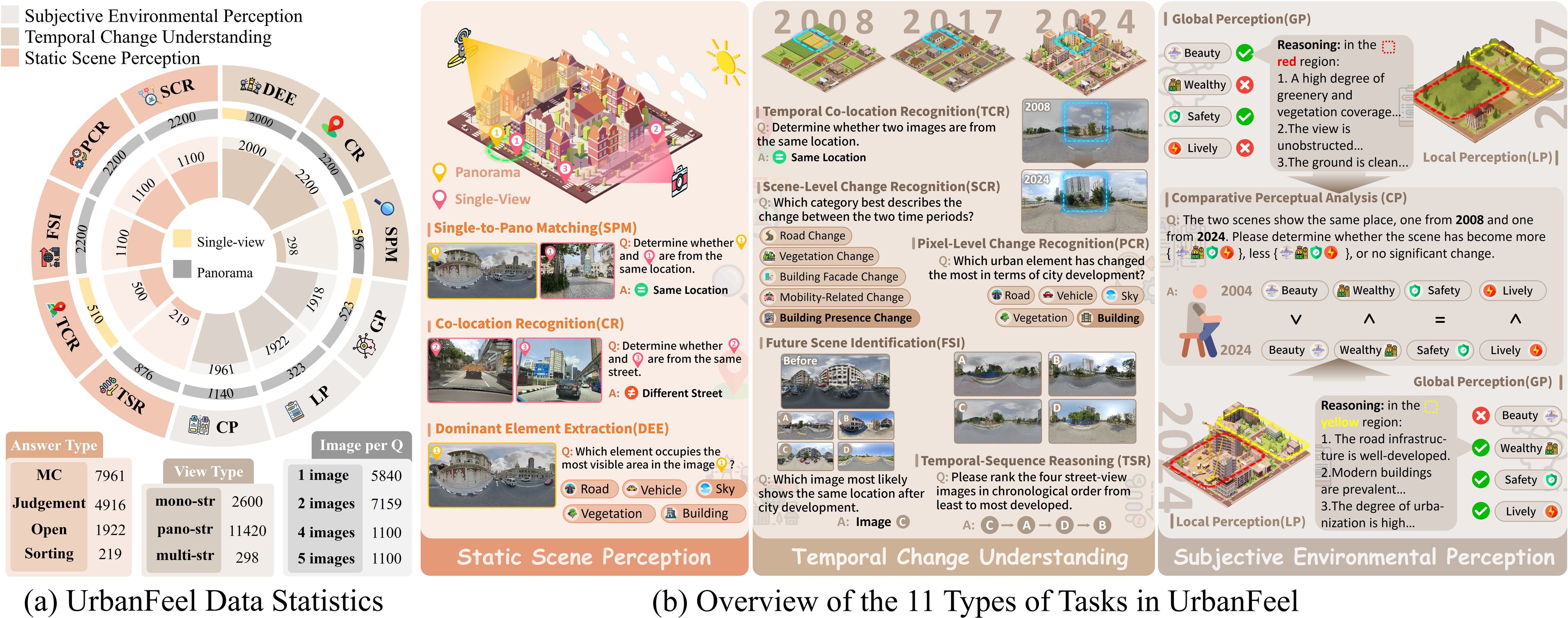}
    \vspace{-6mm}
    \caption{Overview of our UrbanFeel. (a) UrbanFeel comprises over 14.3K carefully designed questions across multiple formats, including multiple choice, binary judgment, open-ended QA, and temporal sorting. (b) UrbanFeel defines 11 sub-tasks spanning 3 cognitive dimensions: static scene perception, temporal change understanding, and subjective environmental perception. }
    \label{fig:task_overview}
    \vspace{-6mm}
\end{figure}

We present \textbf{UrbanFeel}, a comprehensive benchmark designed to evaluate the capabilities of Multimodal Large Language Models (MLLMs) in both physical understanding and subjective perception within the context of urban development. Built upon multi-view and multi-temporal street-view imagery collected from diverse global cities over the past 18 years, UrbanFeel simulates real-world urban evolution by capturing fine-grained visual and perceptual changes. As illustrated in Fig.~\ref{fig:task_overview}(a), UrbanFeel includes four types of question format—binary judgment, multiple-choice, open-ended reasoning, and a novel temporal sorting format—resulting in over 14,300 high-quality QA samples, with approximately 11.0 K for validation and 3.3 K for testing. Detailed statistics and examples can be found in Appendix.

Unlike prior benchmarks that focus primarily on object detection or scene classification within urban imagery, as shown in Fig.\ref{fig:comparison},
UrbanFeel introduces several novel design dimensions. First, it systematically incorporates both single-view and panoramic street images captured in a certain sequence to evaluate models’ ability to capture spatial context across viewpoints. Second, it integrates long-term urban development sequences—spanning more than a decade—to support tasks that require historical reasoning and temporal ordering. Third, UrbanFeel introduces human-centered affective perception tasks, covering four dimensions: beautiful, safety, wealthy, and lively. Each sample is additionally annotated with localized visual evidence, enabling the evaluation of model explainability and alignment with human perceptual cues. UrbanFeel thus offers a challenging and comprehensive evaluation framework for MLLMs in complex urban scenarios, and lays a foundation for future studies on modeling the alignment and divergence between machine and human perception.

\subsection{Benchmark Task}

Guided by a cognitively progressive evaluation framework, we design 11 diverse tasks and construct \textbf{UrbanFeel}, a comprehensive benchmark for modeling urban development perception. As illustrated in Fig.~\ref{fig:task_overview} (b), these tasks span three levels of cognitive depth—\textit{Static Scene Perception}, \textit{Temporal Change Understanding}, and \textit{Subjective Environmental Perception}—enabling a multi-dimensional assessment of MLLMs across recognition, reasoning, and perceptual alignment.

\textbf{Static Scene Perception} focuses on evaluating models’ ability to recognize salient visual elements and spatial consistency in a single time frame. Tasks under this category include identifying dominant visual components in a given image and determining whether a pair of images—single-view and panoramic—depict the same geographic location. This dimension retains some classic scene perception tasks and aims to assess models’ capacity for snapshot-level spatial understanding and contextual matching.

\textbf{Temporal Change Understanding} targets the model's ability to detect, differentiate, and reason about visual changes over time. Beyond identifying structural variations across temporally aligned images, models are required to classify the type of urban evolution (e.g., façade renovation, road maintenance, or vegetation growth) and to perform temporal ordering of multiple images based on perceived development stages. These tasks simulate human-like reasoning about city progression and test the model’s temporal-spatial integration abilities.

\textbf{Subjective Environmental Perception} emphasizes the alignment between MLLMs and human subjective evaluation. We construct affective perception tasks across four dimensions—\textit{beautiful}, \textit{safe}, \textit{wealthy}, and \textit{lively}—and require models not only to produce scalar judgments but also to provide localized visual justifications. In addition, we introduce before–after comparison tasks to examine whether models can detect perceptual shifts in changing environments. This dimension moves beyond objective recognition, probing whether MLLMs can simulate human affective responses in complex visual scenes.

\begin{figure}[t!]
    \centering
    \includegraphics[width=\linewidth]{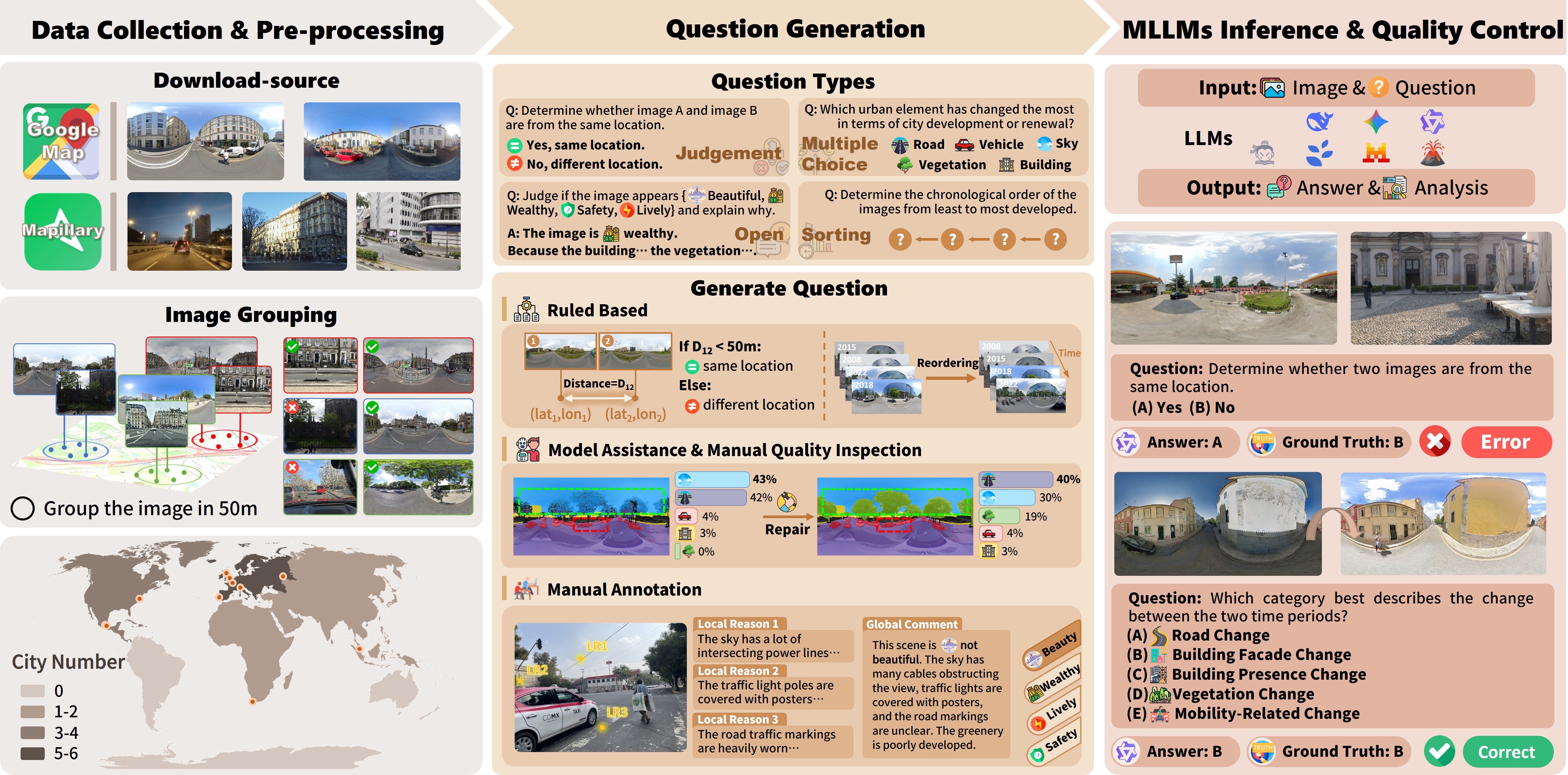}
    \vspace{-6mm}
    \caption{Benchmark construction process of UrbanFeel, including data collection and pre-processing, question generation, and MLLMs inference and quality control.}
    \label{fig:data_management}
    \vspace{-6mm}
\end{figure}

\subsection{Benchmark Curation}

\textbf{Data Collection and Pre-processing.}

As shown in Figure~\ref{fig:data_management}, the UrbanFeel benchmark collects over 4,000 street-view images from 11 cities across four continents via Mapillary and the Google Street View API, covering both single-view and panoramic formats. The selected cities include representative locations from the Global South (e.g., Kuala Lumpur, Tolyatti) and the Global North (e.g., Paris, Washington, D.C.), spanning a temporal range from 2007 to 2024 and capturing diverse stages of urban development.

During the data pre-processing stage, given the lack of precise spatial or temporal ordering in some images, we apply spatiotemporal clustering based on geolocation and timestamps to generate coherent urban evolution sequences. To ensure data quality, we use a pretrained segmentation model along with manual filtering to remove low-quality samples, such as indoor scenes, blurry captures, and heavily occluded images. Additional preprocessing details are provided in the Appendix.

\textbf{Question Generation.}  
UrbanFeel supports four question formats: binary judgment, multiple choice, sorting, and open-ended QA. To efficiently generate diverse question sets, we adopt a hybrid strategy combining rule-based generation, model-assisted prompting, and manual authoring. For instance, tasks like same-place matching and future-view prediction are generated using temporal and spatial metadata. For change-type recognition, we initialize annotations with outputs from general-purpose segmentation models, followed by manual verification and correction. Subjective perception tasks are written entirely by human annotators and span four dimensions: beautiful, safety, wealthy, and lively. Annotators also mark localized visual evidence to support reasoning and explainability evaluation.

\textbf{MLLMs Inference and Quality Control.}  
To ensure annotation accuracy and evaluation reliability, we introduce a multi-stage validation pipeline. During model inference, strict output formatting constraints are enforced. Responses are automatically assessed using a separate language model to compare with human-provided ground truths. For ambiguous or illogical responses, manual review is conducted to reclassify or remove problematic samples. This filtering ensures the final evaluation metrics are robust and reproducible.

\section{Experiments}

\label{sec:Experiments}

\subsection{Evaluated Models}

We evaluate a total of 20 multimodal large models under a zero-shot setting using UrbanFeel, including 2 closed-source models and 18 open-source models. The closed-source models are GPT-4o~\citep{openai_gpt4o_2024} and Gemini 2.5 Pro~\citep{comanici2025gemini25}, accessed via their official APIs. The open-source models cover a diverse set of representative MLLM families, including DeepSeek-VL2~\citep{wu2024deepseek}, InternVL 3~\citep{zhu2025internvl3}, LLaVA~\citep{guo2025llava}, Qwen2.5-VL~\citep{yu2025qwen}, Phi~\citep{abdin2024phi}, Gemma-3~\citep{team2025gemma}, and Idefics3-8B~\citep{laurenccon2024idefics3}, among others. A full list of model versions and configurations is provided in Appendix.

\subsection{Evaluation Protocol}

UrbanFeel includes four question types: binary judgment, multiple choice, sorting, and open-ended QA. Following the evaluation protocol of prior benchmarks such as MMMU~\citep{yue2024mmmu} and UrBench~\citep{zhou2025urbench}, we adopt a hybrid strategy combining exact matching, model verification, and semantic similarity evaluation.

For non-open-ended questions (i.e., judgment, multiple choice, and sorting), we first apply strict string matching—answers are considered correct only if they match the reference label. However, since some models generate verbose responses without clearly selecting an option, we employ an auxiliary language model to assess whether the prediction semantically aligns with the ground truth, ensuring fair evaluation of models that include rationale in their outputs. For open-ended questions, correctness is determined by measuring semantic similarity between the model-generated answer and reference answers. 

\textcolor{black}{For the human baseline, we recruited two independent groups of ten participants each, all with geography-related academic backgrounds (undergraduate, master’s, or doctoral students). One group conducted the annotations, while the other performed the evaluations, ensuring no overlap between the two. More implementation details of evaluation protocol are described in Appendix.}

\subsection{Main Results}

\textbf{Overall Challenge of UrbanFeel.}  
Table~\ref{tab:main_results} summarizes the overall quantitative performance of mainstream Multimodal Large Language Models (MLLMs) on UrbanFeel, revealing the significant challenges posed by our benchmark. While closed-source models such as GPT-4o and Gemini-2.5-Pro demonstrate impressive capabilities on selected tasks, their overall performance remains substantially behind human-level accuracy—particularly in tasks that require compositional reasoning and spatiotemporal understanding. This performance gap is even more pronounced for open-source models, suggesting that current MLLMs still face major limitations in practical applications related to urban development, environmental perception, and city planning.

\begin{table}[!t]
    \centering
    \fontsize{7.3pt}{7.6pt}\selectfont
    \setlength{\tabcolsep}{1.0mm}
    \label{tab:result}
    
    \caption{Quantitative results for 2 closed-source and 18 open-source MLLMs, as well as those for human and random guess across 11 tasks. The overall score is computed across all tasks. The maximum value and the second largest value of model performance in each task are indicated by the \textbf{bold} and \underline{underlined} text, respectively. Task names are abbreviated for brevity.} 
    \vspace{-2mm}
    \begin{tabular}{@{}l|*{3}{>{\centering\arraybackslash}p{0.65cm}}|ccccc|*{3}{>{\centering\arraybackslash}p{1.05cm}}|c@{}}
        \toprule
        \multirow{2}{*}{\textbf{Model}} & \multicolumn{3}{c|}{\textbf{Static Scene Perception}} & \multicolumn{5}{c|}{\textbf{Temporal Change Perception}} & \multicolumn{3}{c|}{\textbf{Subjective Perception Consistency}} & \multirow{2}{*}{\textbf{Overall}}\\
        \cmidrule(lr){2-4} \cmidrule(lr){5-9} \cmidrule(lr){10-12}
         & \textbf{DEE} & \textbf{CR} & \textbf{SPM} & \textbf{TCR} & \textbf{FSI} & \textbf{PCR} & \textbf{TSR} & \textbf{SCR} & \textbf{GP} & \textbf{LP} & \textbf{CP} \\
        \midrule
        DeepSeek-vl2-tiny       & 44.5 & 18.0 & 44.5 & 48.2 & 25.5 & 31.5 & 3.7  & 24.1 & 59.6 & \textbf{53.8} & 43.6 & 36.1 \\
        DeepSeek-vl2            & 59.3 & 29.0 & 43.7 & \underline{94.8} & 53.8 & 37.1 & 8.2  & 38.9 & 65.8 & 40.9 & 33.3 & 45.9 \\
        MiniCPM-V 2.6-8B        & 45.9 & 94.8 & 77.6 & 90.9 & 26.8 & 38.7 & 10.5 & 25.2 & 41.9 & 34.8 & 33.9 & 47.4 \\
        Qwen2.5-vl-3B           & \underline{61.1} & 51.0 & 76.2 & 77.8 & 43.2 & 28.9 & 5.0  & 17.1 & 67.6 & 35.6 & 28.3 & 44.7 \\
        Qwen2.5-vl-7B           & 52.4 & \textbf{98.2} & 75.8 & 85.1 & 43.2 & 33.4 & 10.0 & 43.9 & 56.6 & 33.8 & 38.6 & 51.9 \\
        Qwen2.5-vl-72B          & 60.1 & \underline{97.2} & 66.2 & 87.9 & \underline{90.3} & \textbf{40.9} & 26.0 & 46.0 & 65.7 & 44.6 & 36.9 & \underline{60.2} \\
        LLaVA-1.5-7B            & 26.3 & 51.0 & 49.0 & 88.3 & 25.8 & 24.6 & 3.7  & 17.0 & 59.5 & 39.2 & \textbf{53.2} & 39.8 \\
        LLaVA-v1.6-mistral-7B   & 34.9 & 51.0 & 45.5 & 65.0 & 17.5 & 28.0 & 3.7  & 16.1 & 67.2 & 38.7 & \underline{51.0} & 38.1 \\
        InternVL3-2B            & 51.3 & 41.4 & 65.5 & 66.6 & 19.6 & 38.9 & 1.4  & 15.0 & 68.2 & 40.0 & 25.4 & 39.4 \\
        InternVL3-8B            & 39.8 & 67.4 & 53.8 & 78.4 & 31.8 & 33.5 & 7.8  & 32.2 & \underline{69.7} & 34.1 & 38.8 & 44.3 \\
        Phi-3.5                 & 46.0 & 57.2 & 51.4 & 75.4 & 56.9 & 24.8 & 7.8  & 37.4 & 54.6 & 37.4 & 36.2 & 44.1 \\
        Phi-4                   & 37.7 & 26.2 & 69.7 & 82.9 & 57.4 & 32.1 & 2.3  & 39.9 & \textbf{71.0} & 42.2 & 48.1 & 46.3 \\
        Idefics3-8B             & 47.1 & 56.6 & 60.0 & 53.2 & 22.5 & 39.5 & 3.7  & 15.5 & 61.2 & 34.1 & 49.0 & 40.2 \\
        Mistral-Small-3.1-24B   & 19.6 & 91.6 & 67.9 & 86.6 & 64.3 & 28.9 & 16.0 & 46.7 & 63.3 & 42.4 & 39.9 & 51.6 \\
        Aria                    & \textbf{64.8} & 90.0 & 71.7 & 89.0 & 42.7 & 38.9 & 10.5 & 38.1 & 67.7 & 42.1 & 45.9 & 54.7 \\
        Aya-vision-8b           & 18.8 & 51.0 & 38.6 & 51.5 & 26.2 & 24.6 & 3.2  & 33.3 & 69.6 & 43.8 & 39.5 & 36.4 \\
        Gemma-3-4b              & 42.5 & 81.6 & 64.1 & 59.5 & 25.9 & 39.0 & 8.2  & 36.5 & 53.1 & 46.1 & 26.7 & 43.9 \\
        Gemma-3-27b             & 50.8 & 80.2 & 66.9 & 58.5 & 76.2 & 37.5 & 18.7 & 44.8 & 64.7 & 39.6 & 41.0 & 52.7 \\
        \midrule
        GPT-4o                  & 50.8 & \underline{97.2} & \textbf{79.2} & 89.2 & 74.2 & \underline{40.5} & \underline{38.9} & \underline{49.9} & 60.2 & 37.3 & 36.4 & 59.4 \\
        Gemini-2.5-pro          & \textbf{64.8} & 96.3 & \underline{78.2} & \textbf{95.4} & \textbf{97.6} & 36.5 & \textbf{52.1} & \textbf{56.5} & 67.7 & \underline{49.0} & 30.3 & \textbf{65.9} \\
        
        \midrule
        Human          \quad    & 71.3 & 88.1 & 76.7 & 96.4 & 96.4 & 21.2 & 70.0 & 69.5 & 66.6 & 32.9 & 53.1 & 67.4 \\
        Random         \quad    & 21.4 & 48.6 & 50.3 & 47.6 & 26.3 & 19.5 &  3.9 & 18.8 & 51.0 & 18.2 & 35.2 & 31.5 \\

        \bottomrule
    \end{tabular}
    \label{tab:main_results}
    \vspace{-3mm}
\end{table}

\textbf{Performance Across Task Dimensions.}  
Table~\ref{tab:main_results} further disaggregates model performance across the 11 sub-tasks in UrbanFeel. Most MLLMs exhibit strong capabilities in basic visual recognition tasks; for instance, the majority of models achieve over 60\% accuracy on the Time-Consistent Recognition (TCR) task, which requires only straightforward temporal identification.

However, model performance drops significantly when spatial and temporal reasoning must be integrated. In the TSR task, for example, most models score below 10\% accuracy. Even the best-performing model—Gemini-2.5-Pro—still lags behind human performance by 17.9\% in accuracy, revealing that existing models still struggle with long-range temporal ordering and scene-level alignment across time and space.

Interestingly, in the dimension of subjective environmental perception, many models show strong consistency with human judgment (CP, LP). Their visual justifications—such as cues used to infer aesthetic or safety—often align closely with those identified by human annotators, suggesting early potential for human-aligned perceptual reasoning. However, this alignment weakens substantially when temporal dynamics are introduced. On tasks involving perceptual comparison between before–after scenes, most models exhibit heightened sensitivity to visual changes, often overemphasizing fine-grained variations and diverging from human-level perceptual stability. More subjective analysis cases will be displayed in the Appendix.

\textcolor{black}{To our surprise, in the PCR task with panoramic inputs, MLLMs outperform human evaluators. This is because humans are less sensitive to pixel-level differences caused by panoramic distortions. While evaluators focus on salient foreground changes, mid- or long-range building variations occupy only a small pixel proportion and may be less noticeable than background shifts in sky or road caused by slight camera movements, leading to frequent misjudgments.}

\section{Discussion}

\subsection{Model Performance Across Subjective Perception Dimensions}
\begin{figure}
    \centering
    \includegraphics[width=0.80\linewidth]{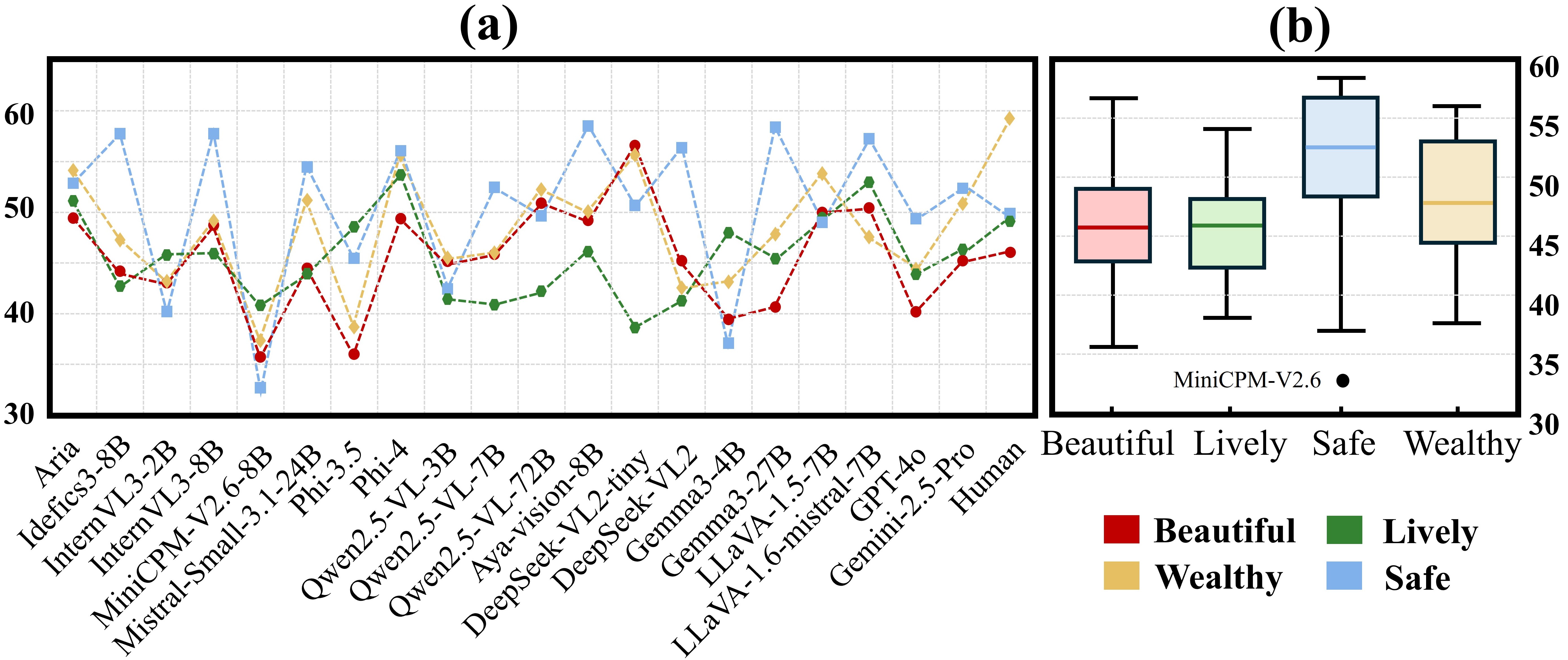}
    \vspace{-2mm}
    \caption{Quantitative comparison of MLLM performance on subjective environment perception. (a) Accuracy across four dimensions. (b) Box plots show model variance, where the horizontal lines in boxes indicate medians; box width indicates consistency.}
    \label{fig:dimension}
    \vspace{-6mm}
\end{figure}

To further investigate model behavior in subjective environmental perception, Figure~\ref{fig:dimension}(a) presents accuracy distributions across four key dimensions. The results show that MLLMs exhibit considerable variation across dimensions. Most models achieve human-comparable or even superior accuracy in dimensions such as \textit{Safe}, \textit{Beautiful}, and \textit{Lively}, suggesting promising potential for aligning with human perceptual judgments in urban scenes.

Among these, \textit{Safe} is the dimension where most models perform best, reaching an average accuracy of 50.6\%. However, this dimension also reveals the largest performance gap between models, indicating substantial inconsistency in safety-related judgments. In contrast, \textit{Lively} displays more stable accuracy across models, despite having a slightly lower average performance, suggesting that models more consistently capture liveliness, likely by relying on broad visual cues such as vehicles and crowds. However, in the \textit{Wealthy} dimension, the models still underperform human evaluators by an average margin of 10.1\%, implying that wealth perception may involve more nuanced or culturally specific visual cues that current models struggle to capture.

The box plot in Figure~\ref{fig:dimension}(b) further supports these observations. Although the \textit{Safe} dimension has the highest median accuracy (52.4\%), it also exhibits the widest interquartile range and largest overall variance, confirming the inconsistent model behavior. Conversely, the \textit{Lively} dimension has the most concentrated distribution, indicating higher inter-model agreement. This consistency suggests that current MLLMs may rely on more universal or easily detectable signals when evaluating liveliness, whereas dimensions like safety and wealth require finer-grained perceptual reasoning or socio-cultural understanding.

\subsection{Does city identity affect MLLMs' subjective environmental judgments?}

\begin{figure}[t!]
    \centering
    \includegraphics[width=0.90\linewidth]{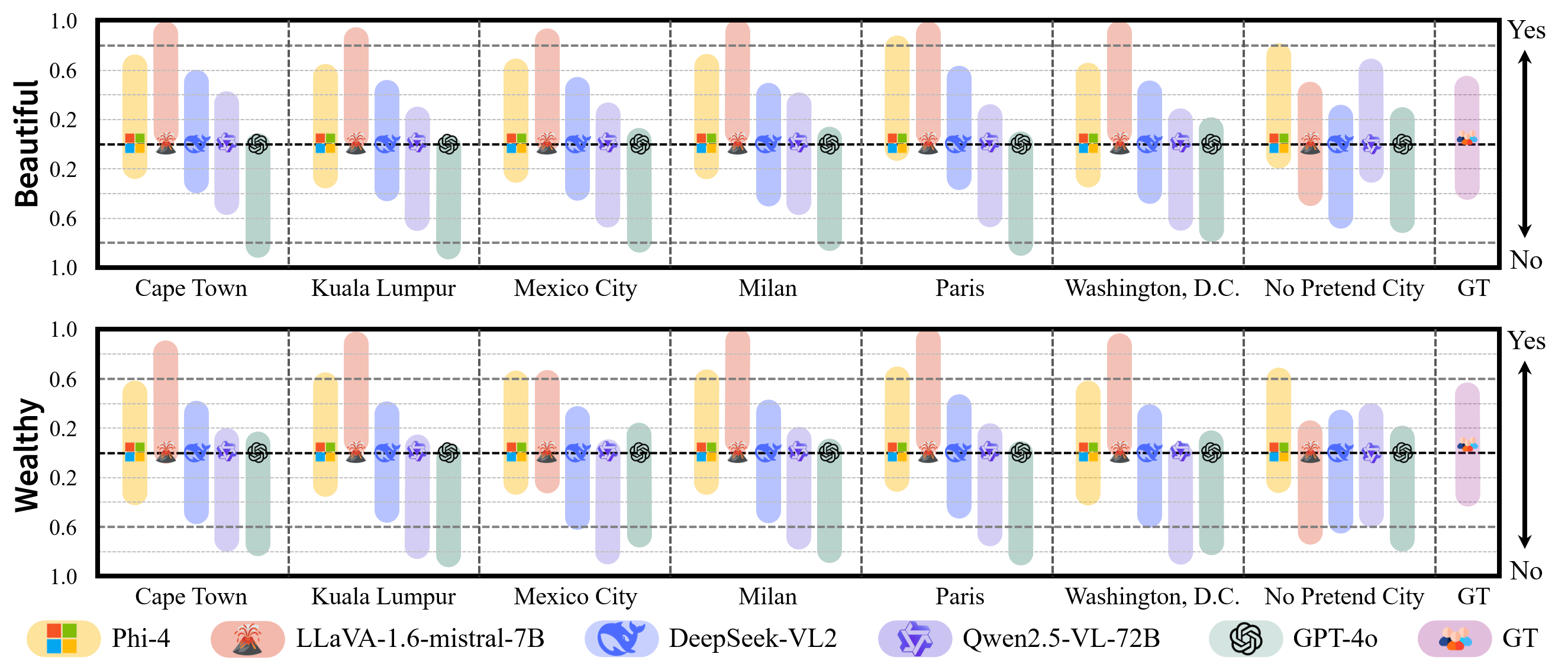}
    \vspace{-2mm}
    \caption{Quantitative comparison of different models under the assumed city identity setting.``Yes" indicates the proportion of positive evaluations made by MLLMs for the given perceptual dimension, while ``No" represents the proportion of negative evaluations. The results show that LLaVA-1.6-mistral-7B and DeepSeek-vl2 yield more positive evaluations across most cities, while Qwen2.5-VL and GPT-4o show a decline under assumed city identity.}

\vspace{-6mm}
    \label{fig:geospatial-bias}
\end{figure}

To examine whether MLLMs exhibit geographic bias in subjective perception, we conducted a "city identity intervention" experiment. We randomly selected 100 street-view images from the GP validation set and assigned each one of six hypothetical city identities (Cape Town, Kuala Lumpur, Mexico City, Milan, Paris, Washington, D.C.), comparing the results to those without any assigned identity (``No Pretend City"). Figure~\ref{fig:geospatial-bias} shows the distributions of positive (e.g., ``beautiful") and negative (e.g., ``not beautiful") judgments under the \textit{Beautiful} and \textit{Wealthy} dimensions. Results for \textit{Lively} and \textit{Safe} are included in the Appendix.

Overall, most models exhibit varying degrees of change in their subjective judgments when city identity is introduced. LLaVA-1.6 and DeepSeek-vl2 tend to produce more positive evaluations across most cities, suggesting a tendency to interpret identity labels favorably. In contrast, Phi-4 demonstrates high stability, indicating greater reliance on image content and robustness to added semantic labels. Notably, GPT-4o and Qwen2.5-VL show a general decline in positive judgments when city identity is provided, implying a more ``cautious” or even ``conservative” evaluation behavior, potentially triggered by the activation of learned stereotypes or expectations.

When comparing ``Global North” and ``Global South” city identities, we observe that\textbf{ a northern identity does not necessarily lead to more favorable evaluations.} Although the average score for northern cities is slightly higher, GPT-4o's perception of wealth for ``Paris'' and ``Milan'' drops significantly—sometimes even below that of cities like ``Cape Town.'' This counterintuitive result may stem from a mismatch between the semantic label and the actual image content; for example, ordinary or aged urban scenes labeled as ``Paris'' may result in greater expectation gaps, prompting the model to generate more negative evaluations. Conversely, GPT-4o's slightly increased positive judgments for ``Mexico City'' may be attributed to positive visual signals such as modern buildings, clean streets, and bright lighting—combined with the lack of strong negative priors associated with the label ``Mexico City'' in the model's pretraining corpus.

\subsection{\textcolor{black}{Do MLLMs Perceive Single-View and Panorama Differently?}}
\begin{wrapfigure}{r}{0.5\textwidth}
  \vspace{-2mm}
  \centering
  \includegraphics[width=\linewidth]{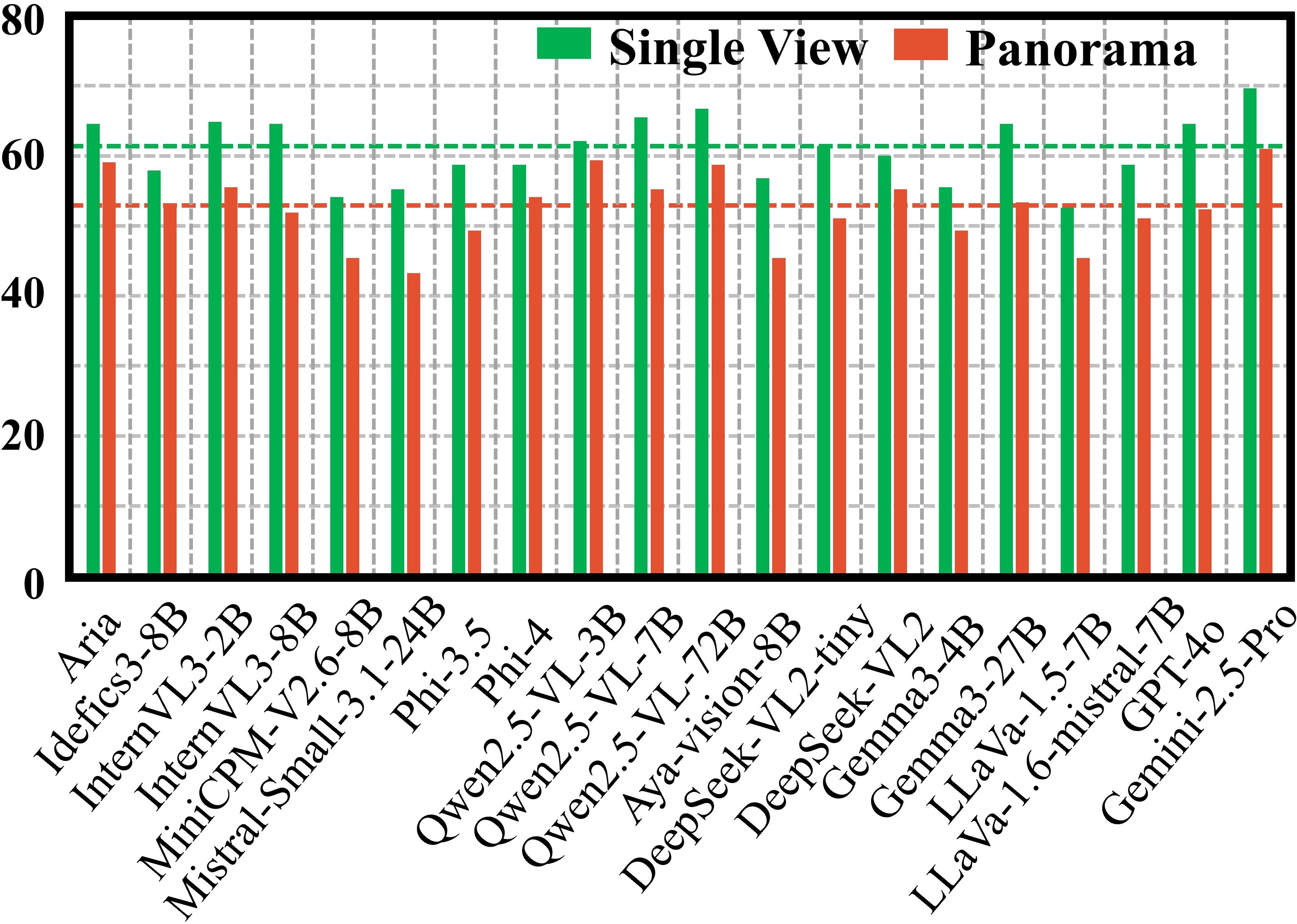}
    \caption{Quantitative results of MLLMs performance from different perspectives.}
    \label{fig:views}
    \vspace{-3mm}
\end{wrapfigure}
To evaluate whether the differences in viewpoint coverage and information organization between single-view and panorama introduce perceptual biases in MLLMs, we compare model performance across the two perspectives. As shown in Figure~\ref{fig:views}, the majority of models consistently perform better on single-view images than on their panorama counterparts. On average, single-view inputs yield an accuracy improvement of 11.7\% over panoramic inputs. Notably, Gemini-2.5-Pro achieves the highest accuracy on single-view images at 69.4\%, closely by Qwen2.5-VL-72B with 64.9\%. In terms of panoramic images, Aria shows the best performance with 55.3\%, while Gemini-2.5-Pro follows closely with with 54.9\%.

This performance gap suggests that although panorama offer greater spatial coverage and denser visual information, their inherent geometric distortions and contextual blending may increase the ``cognitive burden" on MLLMs. It also shows that MLLMs have perspective data imbalance and bias during training process.

\section{Conclusion}
\vspace{-3mm}
In this study, we introduces UrbanFeel, a new benchmark for evaluating the capabilities of Multimodal Large Language Models (MLLMs) in urban development understanding and subjective perception. The benchmark includes over 14.3K questions across 11 tasks, covering static scene perception, temporal change understanding, and subjective environmental perception. It is constructed using single-view and panoramic street-view images from 11 cities, spanning more than 15 years. We evaluate 20 MLLMs and identify key limitations. Current models underperform in tasks requiring joint spatial–temporal reasoning. We also observe geographic bias in subjective perception tasks, where predictions vary with city identity. In addition, models show perceptual inconsistencies across different viewpoints, particularly between single-view and panoramic inputs. We desire UrbanFeel will support future research on perception-aware urban intelligence by highlighting the interplay between temporal urban evolution and subjective human responses.

\bibliography{iclr2026_conference}
\bibliographystyle{iclr2026_conference}

\end{document}